\documentclass[runningheads,a4paper]{llncs}
\usepackage{graphicx}
\usepackage{graphicx}
\usepackage{amsmath}
\usepackage{amssymb}
\usepackage{booktabs}
\usepackage{amsfonts}
\usepackage{adjustbox}
\usepackage{tabularx}
\usepackage{multirow}
\usepackage[normalem]{ulem}
\useunder{\uline}{\ul}{}
\usepackage{authblk}
\usepackage[pagebackref,breaklinks,colorlinks]{hyperref}
\begin{document}
\title{Deepfake Detection via Joint Unsupervised Reconstruction and Supervised Classification
}
\titlerunning{Deepfake Dectection via Joint Reconstruction and Classification}
%
\def\correspondingauthor{\footnote{Corresponding author}}
\author{Bosheng Yan
\orcidID{0000-0001-9591-8727
}\and
Chang-Tsun Li \correspondingauthor
\orcidID{0000-0003-4735-6138}
\and
Xuequan Lu
\orcidID{0000-0003-0959-408X}}
\authorrunning{B. Yan et al.}
%
\institute{Deakin University, Australia \\
\email{\{yanbo,changtsun.li,xuequan.lu\}@deakin.edu.au}}
%

\maketitle           
\begin{abstract}
Deep learning has enabled realistic face manipulation for malicious purposes (i.e., deepfakes), which poses significant concerns over the integrity of the media in circulation. Most existing deep learning techniques for deepfake detection can achieve promising performance in the intra-dataset evaluation setting (i.e., training and testing on the same dataset), but are unable to perform satisfactorily in the inter-dataset evaluation setting (i.e., training on one dataset and testing on another). 
Most of the previous methods use the backbone network to extract global features for making predictions and only employ binary supervision (i.e., indicating whether the training instances are fake or authentic) to train the network. Classification merely based on the learning of global features often leads to weak generalizability to deepfakes attributed to unseen manipulation methods. Previous methods also use a reconstruction process to improve the learned representation. However, they did not consider the merit of combining the reconstruction task with the classification task. In this paper, we introduce a novel approach for deepfake detection, which considers the reconstruction and classification tasks simultaneously to address these problems. This method shares the information learned by one task with the other, each focusing on different aspects, and hence boosts the overall performance. In particular, we design a two-branch Convolutional AutoEncoder (CAE), in which the Convolutional Encoder used to compress the feature map into a latent representation is shared by both branches. The latent representation of the input data is then fed to a simple classifier and the unsupervised reconstruction component simultaneously. Our network is trained end-to-end. Experiments demonstrate that our method achieves state-of-the-art performance on three commonly-used datasets, particularly in the cross-dataset evaluation setting.


\keywords{Deepfake detection  \and Unsupervised  \and Joint learning.}
\end{abstract}
%
%

\section{Introduction}
\label{sec:intro}





Recent deep learning techniques have enabled the generation of realistic face images and the synthesis of tampered videos. The evolution of such techniques allows attackers or malicious users to forge videos/images by replacing the original content with alternative versions, which poses security and privacy threats to the society. 
Therefore, it is of great importance to investigate innovative approaches to detect manipulated faces.

Several existing detection methods for face manipulation are based on visual cues produced by traditional manipulation methods. Yang et al.~\cite{yang2019exposing} proposed a detection approach that analyzes the abnormality of head poses to differentiate the original and deepfakes. Li et al.~\cite{li2018ictu} suggested that the frequency of eye blinking is a good hint to determine if an image/video is manipulated. Li et al.~\cite{li2021exposing} utilized the biological characteristic (eye movement) of the target face to detect deepfake video. Although these methods show promising performance on some benchmark datasets, they are designed for specific visual artifacts caused by the manipulation process. As a result, the performance of their methods drops dramatically when the clues are removed or when the manipulation artifacts are different from the specific artifacts the methods are designed to detect. With the technical evolution of manipulation, a main focus on deepfake detection is deep learning. The method proposed in \cite{afchar2018mesonet,bayar2016deep,feng2020deep} has shown that deep neural networks are effective in distinguishing tampered images from the original ones by capturing discriminative features and local patterns of images. These local patterns are then transformed by the Global Average Pooling into a scalar, resulting in losing the local information. This scalar value is then used for classification in the final stage, meaning the local features do not really contribute to the final decision-making.
Therefore, the detection accuracy of those methods mentioned above degrades significantly when tested on cross datasets because the learned features are strongly related to the specific manipulations used to generate the training data. It also incurs huge time delay to retrain the model. The afore-mentioned observations suggest that  it is crucial to exploit not only global features of the faces~\cite{zhao2021multi,schwarcz2021finding} but also integrate the local face information learned from auxiliary supervision, e.g., unsupervised learning~\cite{kong2022detect} into classification. 

In this paper, we propose a novel method called Joint DeepFake Detector(JDFD) that employ the Convolutional Neural Network (CNN)-based autoencoder for reconstruction and a shallow linear network for classification, respectively.  
In our framework, we use both supervised and unsupervised learning in a complementary fashion. Supervised learning is used to minimise the classification task loss, while unsupervised learning is used for reconstructing the input data without knowing the labels. The reconstruction task uses the shared encoder to learn the representation of the input images' features and helps 
to enhance the generalization ability across different datasets (i.e., cross-dataset evaluation). The framework can be trained end-to-end in a jointly supervised and unsupervised manner. Comprehensive experiments on three datasets allowing (3 intra-dataset evaluations and 6 cross-dataset evaluations) demonstrate the effectiveness of our method, and show that our method achieves state-of-the-art performance on deepfake detection.


In summary, our main contributions are as follows:
\begin{itemize}
  \item We introduce a two-branch Convolutional Autoencoder framework for deepfake detection, which considers both supervised classification and unsupervised reconstruction. Auxiliary supervision is adopted to assist the encoder to extract robust feature representation.
  
  \item We propose to combine the learning of local and global spatial information from the reconstruction and classification tasks, which promotes the ability of mapping the feature representation of the shared encoder. This improves the generalizability of the network to unknown manipulated patterns.
\end{itemize}

\section{Related Work}
\label{sec:relatedwork}

\subsection{Face Manipulation Methods}
Computer vision approaches are widely used for transferring facial expression and features by reconstructing 3D models for both source and target faces, and then exploring the correspondence among face geometry to warp between these two faces.  Thies et al.~\cite{thies2015real} proposed a method to swap facial expression with an RGB-D camera. Face2Face~\cite{thies2016face2face} is a face reenactment system that uses only an RGB camera to manipulate facial expression in real-time. The extended work~\cite{kim2018deep} even transfers more attributes other than expressions such as head position, eye blinking, and rotation from the source individual to a target individual in a video. FaceSwap~\cite{korshunova2017fast} transfers the source individual's features into the target face  while preserving the expressions and head pose. 
Recently, deep learning-based techniques have attracted lots of attention in synthesizing or manipulating face images. Zhmoginov et al.~\cite{zhmoginov2016inverting} proposed a deep learning approach that utilizes the neural network to invert low-dimension face embedding for producing realistic face images. This technique is further extended to the mobile application called FaceApp~\cite{faceapp}, which can selectively edit facial attributes. Another branch of deep learning techniques is GAN-based methods. For example, ProGAN~\cite{karras2017progressive} achieves high-resolution face image synthesis by using progressively growing generator and detector. Although intended for age-oriented face synthesis, the Conditional Discriminator Pool proposed by Wang et al.~\cite{wang2021age} can be easily adapted for deepfake creation. Shao et al.~\cite{shao2019explicit} propose a facial expression transfer framework that specifically swaps the fine-grained expression of two unpaired images while well preserving other natural attributes. In~\cite{fu2021high}, the authors introduce a stage-wise framework with an auto-encoder semi-learning manner. It predicts the image target boundary using pose and expression vector, where the boundary and input image are encoded into a latent space using two encoders. Then the input structure and texture are separated using the LightCNN network, and the concatenation of the boundary and input representation is decoded to produce the target synthesis. They considerably reduce the correlation bias in transferring poses and expressions for high-resolution input.

\subsection{Deepfake Detection Methods}
\label{detection method}
With the increasing magnitude of negative impacts on privacy and information security due to face forgery, researchers are racing to develop effective countermeasures. Some early works
consider artifacts left by face manipulation methods as visual cues, e.g., unnatural eye blinking~\cite{li2018ictu}, abnormality of head pose~\cite{yang2019exposing}, and biological signals from synthesized video~\cite{ciftci2020fakecatcher}. Li et al.~\cite{li2018exposing} proposed a detection method based on exploiting face warping artifacts from the manipulation pipeline. However, all these methods share a common drawback that these artifacts become invalid once the manipulation methods evolve.
As the learning-based methods become popular, a recent study~\cite{rossler2019faceforensics++} proposes the deep learning-based framework that learns the features from the spatial domain and achieves outstanding performance on specific datasets. Zhou et al.~\cite{zhou2017two} proposed a two-stream network to extract steganalysis features for tampered face detection. The work presented in~\cite{yu2022patch} divides the original image into many patches where each patch is a particular facial region. The learned sub-feature maps of these patches are taken as input to the patch-based detector for classification. But their methods only consider a specific manipulation during training, thus, the model is not robust enough when tested on unseen face forgery datasets. Face X-ray~\cite{li2020face} improves the generalization ability to detect fake faces without using fake images generated by existing manipulation methods. However, this method focuses on analyzing the artifacts left in the image by the post-processing step only: the blending step. While the artifacts due to the blending process are useful, other useful artifacts due to the processes at the prior stages in the manipulation pipeline are ignored. Discovering the intrinsic representation between samples from the same class is more useful for detecting general deepfake. By integrating Xception~\cite{chollet2017xception} with a triplet loss in~\cite{feng2020deep}, Feng et al. demonstrated the feasibility of their method in detecting deepfakes. In~\cite{fung2021deepfakeucl}, Feng et al. applies the concept of unsupervised contrastive learning to manipulated face forensics by employing Xception as an encoder network. This method is divided into two training stages: 1) pre-training of the encoder using the original face data and its transformed version as a pair with a contrastive loss, followed by 2) the training of a simple classifier using the output from the encoder. It's worth noting one merit of this method is that they train the encoder without using the ground truth. Although good performance can be observed in intra-data settings, there is still a significant performance gap when their method is applied in cross-data settings. Zhang et al.~\cite{zhang2021deepfake} introduced a self-supervised decoupling network(SDNN) to learn compression and authenticity features. The goal is to normalise the model using various compression rates in order to improve the classification performance of the authenticity classifier. However, the model might not be robust to unseen compression rates since the range of the compression ratios is variable and difficult to predict. In~\cite{li2021frequency}, a frequency-aware discriminative feature learning framework (FDFL) is proposed to minimize intra-class variations of authentic faces while increasing inter-class distance in the embedding space and compensate for the low efficacy of handcrafted features for forgery detection. The single-centre loss (SCL) is presented to drag real face features to the centre and push the manipulated features away. However, the model does not perform well on unseen datasets. 
\begin{figure}[ht]
\centering
\includegraphics[width=12cm]{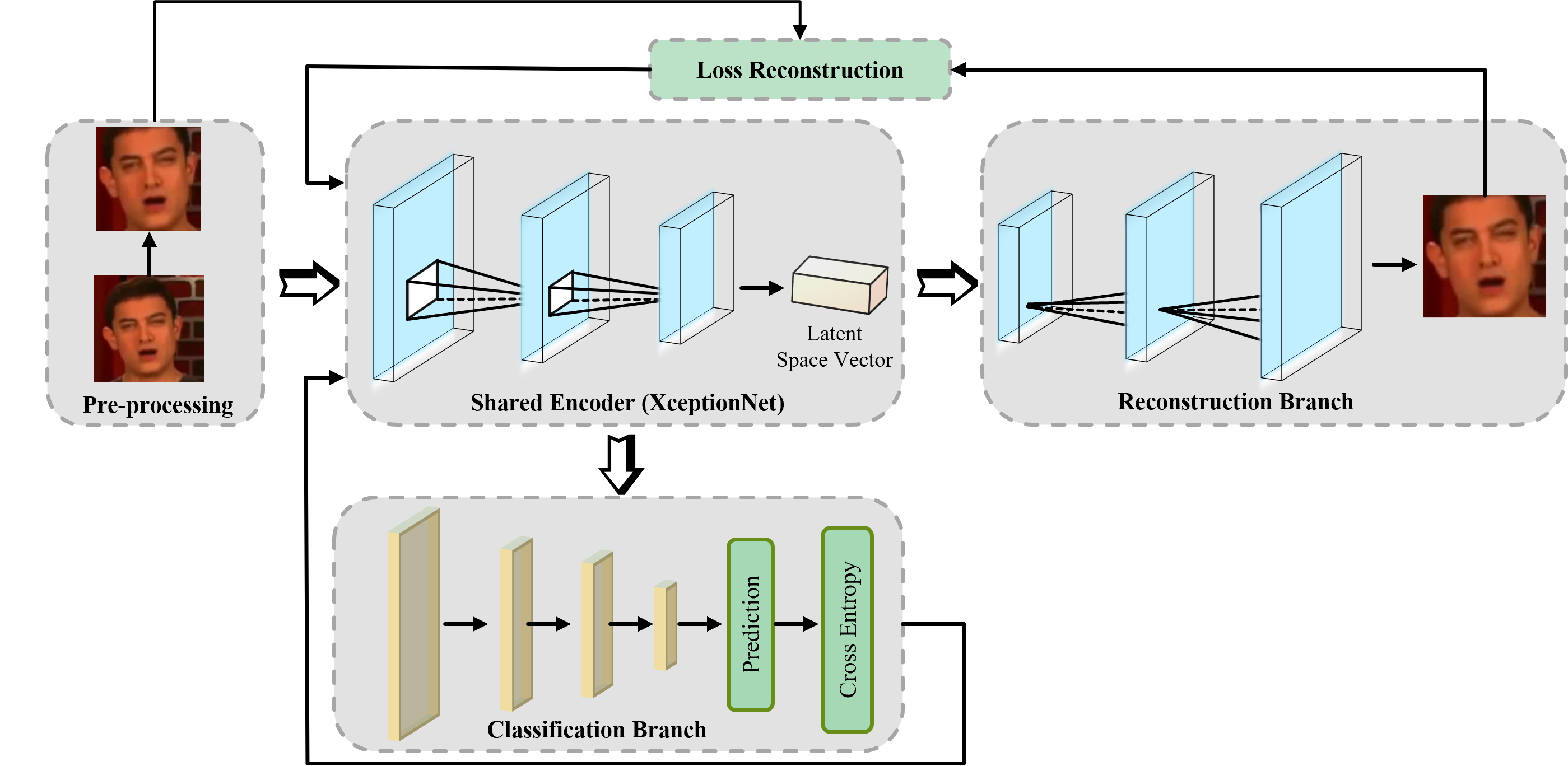}
\caption{Architecture of our proposed framework. Each frame extracted from videos is pre-processed into a face-centred image. The face images are fed into the shared encoder (Xception as backbone) for encoding a latent representation in latent space. The latent vector is fed into the decoder and classifier simultaneously for reconstruction and classification, respectively. }
\label{overview of framework}
\end{figure}


\section{Proposed Method}
\label{methodolog}
Our proposed method produces two outputs including the probabilities and the reconstructed sample of the input. The framework consists of two steps: data pre-processing, and coupled supervised training (classification) and unsupervised training (reconstruction). Previous methods mentioned in Section~\ref{detection method} are dedicated to exploiting the imperfections of a specific manipulation for deepfake detection and neglect the generalization ability in detecting new manipulations. Thus, extracting the discriminative feature representation of the images is essential for differentiating their authenticity. As such, we utilize supervised and unsupervised learning in a complementary fashion to enable the framework to learn more discriminative features which could be explicitly classified by a linear layer network. This approach makes the network robust to unseen face data even if the tampered method is different from the training data. Face-centred images are extracted in the pre-processing step and resized to $299 \times 299$ pixels. Each input is encoded in a high-level latent space through a Convolutional Autoencoder (CAE). We then input the high-level features to the decoder and the classifier simultaneously. Fig.~\ref{overview of framework} shows the overview of our method.

\subsection{Data Pre-processing}
\label{data_pre}

Most of the public face manipulation datasets are stored in the format of video. The face area only takes up a small proportion section in manipulated videos, which would raise difficulties for learning features and predicting probabilities. Since the purpose of deepfake is to swap the face or edit the attributes in the face region, locating the face region of each frame in the video and narrowing down the range of the bounding box to the face centre is necessary. In this paper, we use a popular face extractor Dlib \cite{king2009dlib} to locate the face region in each frame sampled from videos according to 68 facial landmarks and crop the image using a bounding box. To preserve as many tampered traces as possible, the faces are aligned at a centre position to incorporate the spatial information and eliminate the variance of head poses. Finally, the cropped-out faces will go through the transformation procedures (e.g., resize, normalization). The pre-processing allows our framework to work with different resolutions and postures.

\subsection{Two-Branch Autoencoder}
\label{Twobranch}
 
Extracting representations effectively is critical for determining whether face images are real or fake. Different from conventional autoencoder, Convolutional Autoencoder (CAE) applies convolutional layers to learn representation by sharing the same weights among all locations in each input channel, which can preserve spatial locality~\cite{masci2011stacked}. As such, we employ a CAE to extract more distinguishable representations, which are beneficial in separating true and fake images in a high-dimensional space. To force the network to disentangle two different tasks, we feed the latent vector to a shallow linear network to handle the decision while the decoder restores the latent vector to reconstruct the original input by learning the local features of the input. With the combined supervised and unsupervised training, the distinguishable information associated with their classes (real or fake) is separated accordingly. The decoder upsamples the latent vector and reconstructs a realistic sample by learning the local spatial features of input data. Hence, the reconstruction task reinforces the shared encoder to extract robust features. The two-branch autoencoder design ensures the network shares useful information from both the reconstruction and classification tasks, yielding more robust forgery detection.

In this work, the CAE, denoted by $\textit{g}$, consists of two sub-components, namely the encoder $\textit{g}_e$() and the decoder $\textit{g}_d$(). The face dataset, denoted as $\textit{X}$, includes real and fake face images and their corresponding labels ${\{(\textit{x}_i,\textit{y}_i)\}}^N_{i=1}$, where $\textit{x}_i \in \textit{X}$, $N$ is the size, and $\textit{y}_i \in \{0,1\}$. Each training image $\textit{x} \in \mathbb{R}^{w \times h \times 3}$ is taken as input to the encoder for conversion into different low-dimensional feature maps that are projected to the latent vector space to produce the latent vector $\textit{v} \in \mathbb{R}^\textit{z}$, where $\textit{z}$ is the dimension of $\textit{v}$. 
\begin{equation}
    v = g_e(x, W_e),
\end{equation}
where $\textit{W}_e$ denotes parameters for the encoder. Then, the latent vector $\textit{z}$ is sent to the decoder and a linear classifier $\textit{C}\textnormal{()}$, simultaneously as shown in Fig~\ref{overview of framework}. The decoder decodes the latent vector $\textit{v}$ to reconstruct the input image $\hat x \in \mathbb{R}^{w \times h \times 3}$ while the linear classifier estimates the probabilities of $\textit{v}$ being real or fake.
\begin{equation}
    \hat x = g_d(v, W_d), \quad \hat y = \textit{C}(v, W_c),
\end{equation}
where $\textit{W}_d$ and $\textit{W}_c$ are weights of the decoder and classifier, respectively. Note that we only adopt a shallow linear neural network for the classifier because the full-connected layers have a large number of trainable parameters, which are capable of fitting global spatial features rather than relying on the local features generated by convolutional layers. To this end, it is suitable to classify the distributed feature representation encoded in the latent space.
Then the loss information is back-propagated to the encoder to support the extraction of effective representations. 

\subsection{Loss Function}
\label{Loss}

During the training phase, we employ the jointly supervised and unsupervised fashion to optimize the two-branch autoencoder. We enforce the cross-entropy loss $\mathcal{L}_{cro}$ on the predicted authenticity label of the images and the Mean Square Error (MSE) loss $\mathcal{L}_{rec}$ on the images reconstructed from the autoencoder:
\begin{equation}
\label{TotalLoss}    
    \mathcal{L} = \beta_1 \cdot \mathcal{L}_{cro} + \beta_2 \cdot \mathcal{L}_{rec},
\end{equation}
where $\beta _1$ and $\beta _2$ weigh the influence of each loss term. The network is end-to-end trainable with $\mathcal{L}$. 

The cross-entropy loss is to measure the difference between the predicted label $\hat{\textit{y}}_i$ and the ground-truth label $\textit{y}_i$ corresponding to input $\textit{x}_i$. The classifier network is used to predict the probability of each image $\textit{x}_i$: $\sigma(\textit{x}_i) = \frac{exp(\textit{x}_i)}{\sum_{\textit{j=1}}^{N} exp(\textit{x}_j)}$. Let $N$ denote the number of instances. This loss is defined as:
\begin{equation}
    \mathcal{L}_{cro} = \frac{1}{N} \sum_{i}^{N} [\textit{y}_i \text{log}(\sigma(\textit{x}_i)) + (1 - \textit{y}_i)\text{log}(1- \sigma(\textit{x}_i))].
\end{equation}

The reconstruction loss measures the Euclidean distance between the original image $\textit{x}_i$ and the reconstructed image $\hat x_i$, which is produced by $\textit{g}_d(\textit{g}_e(x_i))$: 
\begin{equation}
     \mathcal{L}_{rec} = \frac{1}{N} \sum_{i}^{N} ||\textit{x}_i - \hat{x}_i||^{2}.
\end{equation}



\section{Experiments}
\subsection{Datasets}
\label{database}
We evaluate our proposed method on three different face manipulation datasets: UADFV \cite{li2018exposing}, FaceForensics++ (FF++) \cite{rossler2019faceforensics++}, and Celeb-DF (version2) \cite{Celeb_DF_cvpr20}. The UADFV is a small but commonly used dataset, which includes 49 pristine videos and 49 manipulated videos. Celeb-DF is a relatively large face forgery dataset with the forged images generated with advanced Deepfake algorithms to minimize the perceptibility of the artifacts, such as temporal flickering frames and color inconsistency. FF++ dataset contains 1,000 real videos collected from online archives and their corresponding forged versions generated with four different manipulation methods (DeepFake~\cite{DeepfakesGithub}, Face2Face, FaceSwap~\cite{faceswap}, and NeuralTexture~\cite{thies2019deferred}). For simplicity, we use light compression (c23)~\cite{rossler2019faceforensics++} as our training data. 

The number of images extracted from each dataset is set to 300 frames per video using the pre-processing introduced in Section~\ref{data_pre}. Table \ref{table:image_volume} shows the detailed volume of each dataset we used for training in our experiments.

\begin{table}[]
\caption{Statistics for three datasets: Celeb-DF, FaceForensics++, and UADFV.}
\centering
\renewcommand{\arraystretch}{1.3}
\setlength{\tabcolsep}{5mm}{
\begin{tabular}{lllcllcllcllcll}
\hline
\multicolumn{3}{l}{\multirow{2}{*}{Datasets}} & \multicolumn{6}{c}{Train}                               & \multicolumn{6}{c}{Test}                              \\ \cline{4-15} 
\multicolumn{3}{c}{}                          & \multicolumn{3}{c}{Real}   & \multicolumn{3}{c}{Fake}   & \multicolumn{3}{c}{Real}  & \multicolumn{3}{c}{Fake}  \\ \hline
\multicolumn{3}{l}{Celeb-DF}                  & \multicolumn{3}{c}{152786} & \multicolumn{3}{c}{147901} & \multicolumn{3}{c}{26834} & \multicolumn{3}{c}{26108} \\ 
\multicolumn{3}{l}{FaceForensics++}           & \multicolumn{3}{c}{117218} & \multicolumn{3}{c}{124429} & \multicolumn{3}{c}{20795} & \multicolumn{3}{c}{21995} \\ 
\multicolumn{3}{l}{UADFV}                     & \multicolumn{3}{c}{9881}   & \multicolumn{3}{c}{9886}   & \multicolumn{3}{c}{1740}  & \multicolumn{3}{c}{1723}  \\ \hline
\end{tabular}
}
\vspace*{-0.6\baselineskip}
\label{table:image_volume}
\end{table}

\begin{figure}[ht]
\includegraphics[scale=0.4]{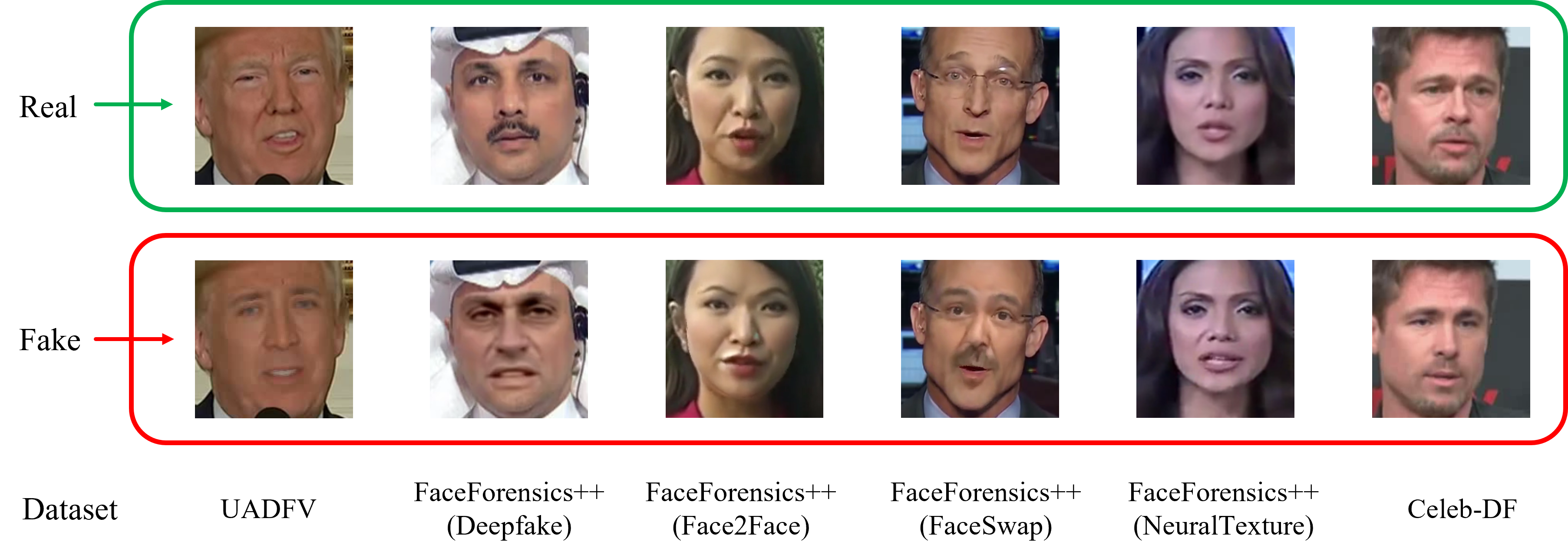}
\caption{Real and fake face samples after pre-processing from three datasets (UADFV, FaceForensics++, and Celeb-DF).}
\label{samples}
\end{figure}

We show some examples of pre-processed images from UADFV, Celeb-DF, and each category of the FF++ dataset in Fig.~\ref{samples}. The listed images show that they are at the similar visual quality level (e.g., brightness, image resolution). However, the fake images in UADFV are easier to be distinguished by human eyes because of the blending boundary discrepancies, unnatural facial expressions, and color inconsistency.

\subsection{Implementation Details}
\label{implementation detail}
For image pre-processing, we use Dlib \cite{king2009dlib} as explained in Section~\ref{data_pre} to detect face regions, crop the faces, and resize the images to $299$ $\times$ $299$. In this work, we adopt Xception~\cite{chollet2017xception} as the backbone of the shared encoder. The structure of the decoder and shallow linear classifier are pre-defined. Following each deconvolutional layer are a batch normalization~\cite{ioffe2015batch} and a rectified linear unit (ReLU)~\cite{nair2010rectified} except the last deconvolutional layer. We resize the reconstructed images to 299 $\times$ 299 using interpolation with the bi-linear mode. For the classifier network, we apply ReLU after each fully connected layer excluding the last dense layer. The network is trained with the SGD optimizer with the learning rates of the convolutional autoencoder and the classification network set to 0.005 and 0.0004, respectively. We also add a scheduler on our optimizer with step size 5 and $80\%$  descending rate. The batch size is set to $4$ in our experiments. 
We empirically set $\beta_1=0.8$ and $\beta_2=0.2$.

\subsection{Comparison with Previous Methods}
In this section, we compare our proposed framework, JDFD with state-of-the-art face forgery detection methods. We provide intra-dataset and cross-dataset evaluations on UADFV, FF++ and Celeb-DF. AUC (area under receiver operating characteristic curve) is adopted as the metric to evaluate the performance.


\begin{table}[]
\centering
\caption{AUC (\%) of Intra-dataset evaluation setting and Cross-dataset setting on three datasets (FaceForensics++, UADFV and Celeb-DF). Best results for intra-dataset setting are underlined, and best results for cross-dataset setting are in bold. }
\renewcommand{\arraystretch}{1.1}
\setlength{\tabcolsep}{1.45mm}{
\begin{tabular}{lcccc}
\hline
Methods        & \multicolumn{1}{l}{Train data} & \multicolumn{1}{l}{FF++} & \multicolumn{1}{l}{UADFV} & \multicolumn{1}{l}{Celeb-DF} \\ \hline
Xception\cite{rossler2019faceforensics++}       & FF++                           & 99.7                     & 80.4                      & 48.2                         \\
Capsule\cite{nguyen2019capsule}        & FF++                           & 96.6                     & 61.3                      & 57.5                         \\
Xception+Tri.\cite{feng2020deep}  & FF++                           & {\ul99.9}                     & 74.3                      & 61.7                         \\
Xception\cite{Celeb_DF_cvpr20}       & UADFV                          & -                        & 96.8                      & 52.2                         \\
Xception+Reg.\cite{dang2020detection}  & UADFV                          & -                        & 98.4                      & 57.1                         \\
Xception+Tri.\cite{feng2020deep}  & UADFV                          & 61.3                     & {\ul 99.9}                      & 60.0                         \\
Headpose\cite{yang2019exposing}       & UADFV                          & 47.3                     & 89.0                      & 54.6                         \\
FWA\cite{li2018exposing}            & UADFV                          & \textbf{80.1}                     & 97.4                      & 56.9                         \\
Xception+Tri.
\cite{feng2020deep}& Celeb-DF                       & 60.2                     & \textbf{88.9}                      & {\ul99.9}                         \\ \hline
JDFD  & FF++                           & 98.2                     & \textbf{92.9}             & \textbf{78.0}                \\
JDFD  & UADFV                          & 61.0                     & {\ul 99.9}                & \textbf{64.9}                \\
JDFD  & Celeb-DF                       & \textbf{61.1}                     & 88.4                      & 97.1                         \\ \hline
\end{tabular}}
\label{STOAcomparison}
\end{table}

From Table~\ref{STOAcomparison}, we can observe the AUC values by different approaches. Regarding the evaluation result from the model trained on FF++, our method is $1.5\%$ weaker than Xception and $1.7\%$ weaker than Xception+Tri when tested on the FF++ dataset. 
For the cross-dataset setting, we observe a dramatic performance improvement of our method when tested on UADFV and Celeb-DF (e.g., outperforming Xception by $12.5\%$ on UADFV and Xception+Tri by $16.3\%$ on Celeb-DF). Because the decoder models the distribution of the latent vector corresponding to each input data, the reconstruction loss reinforces the learning of input data's inherent features of the category it belongs to. To this end, our joint supervised and unsupervised learning enables the encoder to learn robust representations, allowing it to generalize better to unknown forged patterns. Celeb-DF is a challenging Deepfake dataset among these three in terms of quality, quantity, and diversity. For results from the model trained on UADFV, we reach the state-of-the-art performance in the intra-dataset setting. Also, our method outperforms the best-performing method of others by $4.9\%$ on Celeb-DF. However, FWA \cite{li2018exposing} achieves excellent AUC on the FF++ dataset which is $19\%$ higher than our method. This is similar to \cite{feng2020deep}, which is because their method \cite{li2018exposing} focuses on specific face discrepancies due to the affine transformations employed by various deepfake generation methods used to generate some benchmarking datasets (e.g., UADFV, FF++). To enhance our method's generalization power, we do not design it to detect specific manipulations but to learn both local batch patterns and global spatial information. Therefore, it might be slightly weaker than these methods in some specific cases. Given the fact that the deepfake generation algorithms evolve, countermeasures designed for detecting specific manipulations will soon become unfit for purposes in the near future. Celeb-DF is a more challenging dataset for deepfake detection. When trained on Celeb-DF, the result of our method is still competitive when compared to Xception+Tri \cite{feng2020deep}. With the performance on the cross-dataset setting, our method is 0.9\% higher on FF++ and $0.5\%$ weaker on UADFV. As our method considers the learned local features into classification rather than just determining global features, the performance is better on FF++ (containing four different manipulation methods) demonstrating that our method improved the generalizability of the model. In summary, our method generally achieves better results on the cross-dataset evaluation setting (e.g., it is the best performer in 4 of the 6 tests).

\begin{figure}[ht]
\centering
\includegraphics[scale=1]{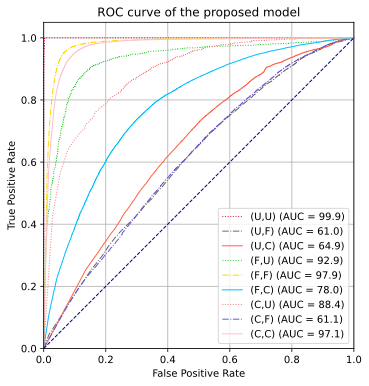}

\caption{Receiver Operating Characteristic (ROC) curves of the our method for intra-data and cross-data settings. U, F, and C in the figure represent UADFV, FaceForensics++, and Celeb-DF, respectively. The first bracket specifies the training and testing datasets. For instance, (U, F) indicates that the model is trained on UADFV and tested on FaceForensics++.}
\label{auc}
\end{figure}

The ROC curves of the corresponding AUCs of our method are also shown in Fig.~\ref{auc}. The diagonal line indicates the baseline result from random classification (50\% in binary classification). Thus, the curve further away from the diagonal line from above indicates better performance. 

\subsection{Ablation Study}
In this section, we show two ablation studies: the effect of unsupervised learning, and the effect of data augmentation from other datasets.

\subsubsection{Effect of unsupervised learning}
To evaluate the contributions of unsupervised learning, we set up a baseline model to compare with our proposed framework. The baseline model has the same architecture but without the decoder.
In Table~\ref{BaselineComparison}, we report the results of these two models. With the unsupervised learning branch, our model has achieved obviously better performance on the intra-dataset evaluation of UADFV, FF++, and Celeb-DF, with  0.4\%, 0.8\%, and 1.5\% improvement, respectively. The generalization capability to new forms of manipulation becomes stronger, as seen by the results of our method outperforming the baseline model on all the cross-dataset settings. When trained on UADFV, the AUCs of our method are 2.6 \% and 1.5\% higher when tested on FF++ and Celeb-DF, respectively. There is a significant increase (12.2\%) when trained on FF++ and tested on UADFV, and a moderate increase (2.6\%) when trained on FF++ and tested on Celeb-DF. The performance of our method increases to 88.4\% when trained on Celeb-DF and tested on UADFV, which is 5.3\% higher than the results of the baseline model. So, the proposed combination of supervised learning and unsupervised learning enables the shared encoder to not only extract the local feature by sliding the window but also learn the relationship between samples by mining the inherent representation of the data. Then the compressed latent vector is beneficial to determine the authenticity of the input sample.

\begin{table}[]
\centering
\caption{AUC (\%) comparison with the baseline model on UADFV, FaceForensics (FF++) and Celeb-DF. }
\setlength{\tabcolsep}{1.45mm}{
\renewcommand{\arraystretch}{1.2}
\begin{tabular}{llllllclclcl}
\hline
\multicolumn{3}{l}{Train data}                & \multicolumn{3}{l}{Method}        & \multicolumn{2}{c}{UADFV} & \multicolumn{2}{c}{FF++} & \multicolumn{2}{c}{Celeb-DF} \\ \hline
\multicolumn{3}{l}{\multirow{2}{*}{UADFV}}    & \multicolumn{3}{l}{Baseline}      & \multicolumn{2}{c}{99.5}  & \multicolumn{2}{c}{58.4} & \multicolumn{2}{c}{63.7}     \\
\multicolumn{3}{l}{}                          & \multicolumn{3}{l}{JointDeepfake} & \multicolumn{2}{c}{99.9}  & \multicolumn{2}{c}{61.0} & \multicolumn{2}{c}{65.2}     \\ \hline
\multicolumn{3}{l}{\multirow{2}{*}{FF++}}     & \multicolumn{3}{l}{Baseline}      & \multicolumn{2}{c}{80.7}  & \multicolumn{2}{c}{97.4} & \multicolumn{2}{c}{75.4}     \\  
\multicolumn{3}{l}{}                          & \multicolumn{3}{l}{JointDeepfake} & \multicolumn{2}{c}{92.9}  & \multicolumn{2}{c}{98.2} & \multicolumn{2}{c}{78.0}     \\ \hline
\multicolumn{3}{l}{\multirow{2}{*}{Celeb-DF}} & \multicolumn{3}{l}{Baseline}      & \multicolumn{2}{c}{83.0}  & \multicolumn{2}{c}{59.7} & \multicolumn{2}{c}{95.6}     \\  
\multicolumn{3}{l}{}                          & \multicolumn{3}{l}{JointDeepfake} & \multicolumn{2}{c}{88.4}  & \multicolumn{2}{c}{61.1} & \multicolumn{2}{c}{97.1}     \\ \hline
\end{tabular}}

\label{BaselineComparison}
\end{table}

\subsubsection{Data augmentation from other datasets}
\begin{figure}[ht]
\centering
\includegraphics[scale=1]{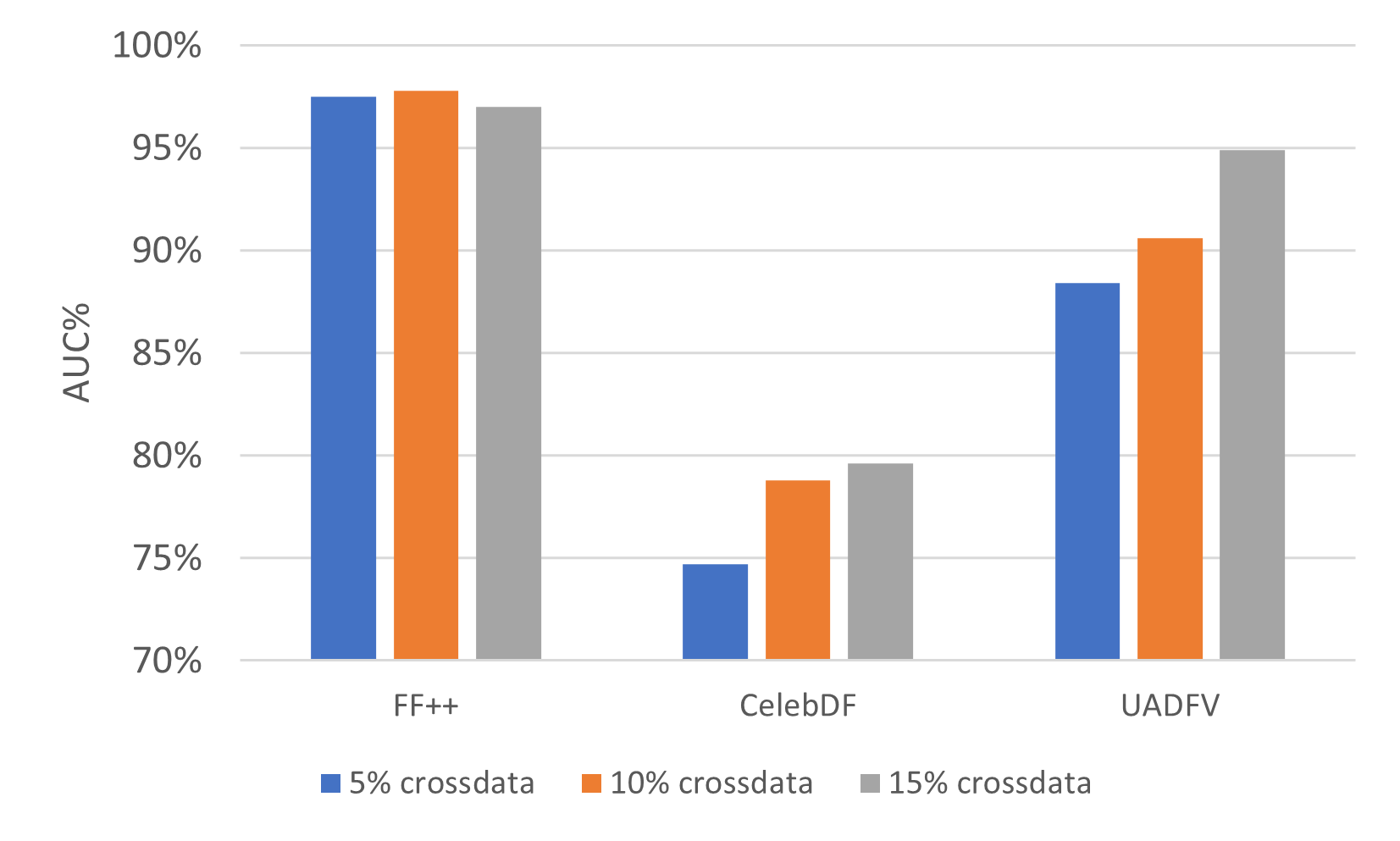}
\caption{AUC (\%) of three different augmented data ratios (5\%, 10\%, 15\%, labels are assumed to be unknown) from UADFV and Celeb-DF datasets. The model is trained on the FF++ dataset.}
\label{data_augmentation}
\end{figure}

It is interesting to evaluate the generalization capability of the proposed method when new data is added. 
We add more images from the other two datasets other than the current dataset used for training. Note that the labels of the added images are assumed to be unknown during training. 
This study involves three schemes. $\textit{1)}$ We randomly choose 5$\%$ images from each of the other two datasets (e.g., around 15,000 images from FF++).  $\textit{2)}$ We increase the augmented foreign data to 10$\%$. $\textit{3)}$ The augmented foreign data is increased to 15$\%$. The results are shown in Fig.~\ref{data_augmentation}. In each augmented data group from each dataset, the ratio of real/fake images is around 50\%. 
We train the model on FF++ with the configuration demonstrated in Section \ref{implementation detail}. The results demonstrate that the AUC generally increases with respect to the amount of augmented data. When 15\% foreign data is added to our framework for training, the performance on UADFV improves by 6.5$\%$ when compared to the case of 5$\%$ cross data. Similarly, the performance for Celeb-DF improves significantly to 79.6\%, which is 4.9\% higher. When tested on the FF++ itself, the result of adding 15\% foreign data drops by around 0.8\% compared with the best result achieved in this study. This is understandable as adding foreign data will affect the training due to the added diversity introduced into FF++, but will naturally enhance the cross-dataset evaluations due to the augmentation of foreign data itself.

\section{Conclusion}
In this paper, we introduce a two-branch autoencoder with jointly unsupervised and supervised learning for deepfake detection. The shared encoder is optimized by receiving the back-propagated information from the reconstruction task and classification task, which enables the encoder to learn the representations effectively. Experimental results demonstrate that joint supervised and unsupervised learning can effectively promote the encoder's efficacy in projecting discriminative features onto the latent space and differentiate the representations between real and forgery samples. These latent vectors are fed into the shallow linear network for prediction. It is able to perform well in deepfake detection, in particular in the inter-dataset evaluation settings. 
Future work will focus on developing an attention module in our framework to locate different forgery regions, which helps to disentangle more potential clues in the latent space for the detector and improve the ability of the framework to cope with seen and unseen face manipulation.

%
%
%
%
{\small
\bibliographystyle{splncs04}
\bibliography{samplepaper}
}





\end{document}